\documentclass{article}
\usepackage{spconf,amsmath, graphics}
\usepackage[fleqn]{nccmath}

\usepackage[dvipsnames]{xcolor}
\usepackage{dsfont}
\usepackage{txfonts}
\usepackage{booktabs}
\usepackage[flushleft]{threeparttable}
\usepackage{xcolor}
\usepackage[textsize=tiny]{todonotes}
\usepackage[citecolor=blue,colorlinks=true]{hyperref}
\usepackage{cleveref}

\newcommand{\myworries}[1]{\textcolor{black}{#1}}

\title{Optimal transport distances for directed, weighted graphs: a case study with cell-cell communication networks}
\name{James S. Nagai$^{a}$ \qquad Ivan G. Costa$^{a}$\sthanks{JSN and IGC acknowledge funding from the Bundesministerium für Bildung und Forschung (BMBF e:Med Consortia Fibromap and CompLS Consortia Graphs4Patients).} \qquad Michael T. Schaub$^{b}$\sthanks{MTS acknowledges funding by the Ministry of Culture and Science (MKW) of the German State of North Rhine-Westphalia ("NRW Rückkehrprogramm")"} }
\address{$^{a}$ Institute for Computational Genomics, RWTH Aachen Medical Faculty, Germany \\
$^{b}$Department of Computer Science, RWTH Aachen University, Germany}
\begin{document}
\ninept
\maketitle
\begin{abstract}



Comparing graphs by means of optimal transport has recently gained significant attention, as the distances induced by optimal transport provide both a principled metric between graphs as well as an interpretable description of the associated changes between graphs in terms of a transport plan.  
As the lack of symmetry introduces challenges in the typically considered formulations,  optimal transport distances for graphs have mostly been developed for undirected graphs.
Here\footnote{\url{https://scaccordion.readthedocs.io/en/latest/}}
, we propose two distance measures to compare directed graphs based on variants of optimal transport(OT): (i) an earth movers distance (Wasserstein) and (ii) a Gromov-Wasserstein (GW) distance. 
We evaluate these two distances and discuss their relative performance for both simulated graph data and real-world directed cell-cell communication graphs, inferred from single-cell RNA-seq data.
\end{abstract}

\section{Introduction}

Exploring the similarities between graphs is a crucial primitive for comparing patterns in complex networks. 
A large variety of distance measures between graphs exist, often based on comparing features derived from (algebraic representations) of the graph structures. 
Examples include distance measures based on spectral features, the computation of certain subgraph statistics such as graphlets~\cite{Malod:2015} or graph kernels~\cite{vishwanathan2010graph}.
While the comparison of such graph features is powerful, it often does elucidate where precisely one graph differs from another, or how two graphs align, respectively.

Optimal transport(OT) based graph distances~\cite{Maretic:2019,xu:2019}, which have risen to prominence recently and address these challenges.
In a nutshell, OT-based graph distances are associated with a certain probability distribution for each graph. 
Two graphs can then be compared by finding a transport plan (a mapping) between those two probability distributions with the minimal transport cost~\cite{Peyre:2019}.
The transport plan associated with the minimal cost can then be used as an interpretable and robust alignment of the two graphs considered, highlighting the changes between the graphs relevant to the computed distance.

To date, most OT-based methods for measuring network similarities have been proposed for undirected graphs.
In many applications, however, we are interested in comparing directed graphs.
Yet, extending OT-based distances to directed graphs is not simple, as a \emph{symmetric} distance metric, typically derived from the distances between nodes in the graph, is required within the cost function(s) typically employed within OT.
To address this problem, we consider two node-to-node distances, which have been developed for directed graphs: the Generalized Effective Resistance (GRD)~\cite{young2:2015} and Markov chain hitting time (HTD)~\cite{Boyd:2021}.
With these distance measures, we can compute OT-based graph distances even for directed graphs.
Specifically, we explore the use of these metrics for both Wasserstein (Earth Mover) and Gromov-Wasserstein based OT distances for graphs~\cite{Peyre:2019}. 

Having established our OT-based distances for directed graphs, we evaluate their relative utility in the context of clustering cell-cell communication networks which arise in the study of single-cell sequencing data.
These networks are intrinsically directional and provide a challenging test case, as technical artifacts such as data dropouts (e.g., missing connections and entities), outliers, and noise are often present in those~\cite{Nagai:2021}.

\noindent\textbf{Contributions} 
(i) We present and evaluate two OT formulations (Wasserstein and Gromov Wasserstein) to compare directed graphs based on two node-to-node distances for directed graphs, namely, the Markov chain hitting time ~\cite{Boyd:2021} and the Generalized Effective Resistance~\cite{young2:2015} 
(ii) We evaluate proposed approaches with simulated directed stochastic block models and on a case study of patient cell-cell communication networks.

\noindent\textbf{Outline}
The remainder of the paper is structured as follows.
We briefly outline some related work in~\Cref{sec:related_work}.
In~\Cref{sec:methods}, we then discuss two node-to-node distance measures for directed graphs, that we employ to formulate two optimal transport distances between directed graphs.
We illustrate the utility of these different formulations in~\Cref{sec:numerical_experiments}, in which we provide some numerical illustrations of these distances for synthetic and real world data.
We close with a short discussion outlining future work.

\section{Related work}
\label{sec:related_work}

\textbf{Optimal transport distances between graphs.}
One of the earliest OT-based distances for comparing undirected graphs considers vectorial embeddings of the graph adjacency matrices and then computes graph similarities using a Wasserstein metric (earth movers' distance) between the embeddings~\cite{Nikolentzos:2017}. 
A different perspective is taken by Graph Optimal Transport (GOT)~\cite{Maretic:2019}. 
Here, the pseudoinverse of the graph Laplacian is treated as the covariance of a multivariate Gaussian distribution, and a multivariate Gaussian OT problem~\cite{Takatsu:2010} is then solved to define the distance between two graphs.

In the context of directed graph, far fewer distances between graphs exists.
Recently, Silva and collaborators~\cite{Silva:2023} have explored the use of Wasserstein distances between distributions of directed graphlets to compare directed graphs. 
Other OT formulations for the comparison of directed graphs are based on Gromov-Wasserstein formulations (GWOT)~\cite{Memoli:2011,Chowdhury:2019}, where the node-to-node distances within the graph are transported instead of the node embeddings.

\noindent\textbf{Node-to-node distance metrics for directed graphs}
When considering undirected graphs, the most commonly adopted node distance measures are the shortest-path distance, or the resistance distance~\cite{Randic:1993}.
For directed graphs, defining a distance measure between nodes becomes a nontrivial task, however, due to the non-symmetric nature of the graph. 
In the following, we leverage two formulations for a directed distance measure between nodes in a graph to define an optimal transport distance between graphs.
By considering the Markov Chain associated with a directed weighted graph, Young and collaborators~\cite{young2:2015} extended the resistance distance to directed and weighted graphs. 
A related formulation has been provided by \cite{Boyd:2021}, which introduced a node-to-node pseudo-metric based on Markov Chain Hitting times. 

\section{Methods}
\label{sec:methods}

In this section, we define two distance measures between \emph{directed, weighted graphs} based on optimal transport theory.
We consider a setup in which we are confronted with a set of directed weighted graphs $\lbrace\mathcal{G}^{1},..., \mathcal{G}^{p}\rbrace$ where each $\mathcal{G}^i$ denotes a digraph $\mathcal{G}^{i}=(\mathcal{V}^{i},\mathcal{E}^{i},w^{i})$ consisting of a node set $\mathcal{V}$ a set of directed edges $\mathcal{E}^i \subset \mathcal V \times \mathcal V$ and an associated non-negative weight function $w^i:\mathcal E \rightarrow \mathbb{R}_\geq0$.

\myworries{Our task is now to define} a distance function $d(\mathcal{G}^{k},\mathcal{G}^{l})$  between any two graphs by means of optimal transport.
To obtain a well-defined optimal transport problem, we define a geometry according to which the transport cost can be computed.
In the following, we describe two ways to obtain such a geometry from directed weighted graphs, specifically from node-to-node distance metrics defined for directed graphs.
We then employ these two node-to-node distances within two different optimal transport formulations to compare directed weighted graphs, yielding all up 4 possible distance measures.

\subsection{Node-to-Node distances for directed graphs}
\subsubsection{Generalized Effective Resistance Distance (GRD)}
The effective resistance or resistance distance is a well known distance metric for nodes within undirected networks \cite{Klein:1993,ghosh2008minimizing}. 
To derive the resistance distance for undirected graphs, we consider a graph as resistor network, where the weight of each edge corresponds to a conductance. 
The resistance distance between two nodes $i$ and $j$ in the graph is then equal to the potential difference that is induced by injecting a unit current between two nodes~\cite{Klein:1993,ghosh2008minimizing}.

Formally, let $\textbf{A} \in \mathbb{R}^{N\times N}$ be the adjacency matrix of a graph with $N$ nodes, with entries $A_{ij} = w_{ij}$ equal to the weight $w_{ij}$ of the edge from node $i$ to $j$, and $A_{ij}=0$ otherwise.
The Laplacian of the graph is then given as $\textbf{L}=\textbf{D}-\textbf{A}$, where $D=\text{diag}(\mathbf{A1})$ is the diagonal matrix of (weighted) node degrees.
The (square root of the) resistance distance between node $i$ and $j$ can then be computed as~\cite{Klein:1993,ghosh2008minimizing}: 
\begin{equation}
r_{ij} = \sqrt{(e_i-e_j)^\top \textbf{L}^{\dagger} (e_i-e_j)}, 
\end{equation}
where $\textbf{L}^\dagger$ denotes the Moore-Penrose pseudoinverse of the Laplacian, and $e_i, e_j$ are the $N$ dimensional indicator vectors associated to node $i$ and node $j$, respectively. Note that \myworries{$e_i$ is a vector with 1 in the $i$-th position and 0 otherwise.}

To generalize this notion of effective resistance to directed graphs, we now consider the following formulation, due to Young et al.~\cite{young1:2015,young2:2015}.
Let  $\textbf{Q} \in \mathbb{R}^{N-1 \times N}$ be a (grounding) matrix that 
satisfies:
\begin{ceqn}
\begin{align}
\mathbf{Q1} _{N} = 0 \qquad \textbf{QQ}^\top=\textbf{I}_{N-1} \qquad \textbf{Q}^\top \textbf{Q}= \textbf{I}- \mathbf{11}^\top/N.
\end{align}
\end{ceqn}

A \emph{grounded} Laplacian matrix for the graph can now be computed via $Q$ as follows:
\begin{ceqn}
\begin{align}
\widetilde{\textbf{L}} = \textbf{QLQ}^\top.
\end{align}
\end{ceqn}
Using the grounded Laplacian, the (square root of) the resistance distance can now be alternatively written as $r_{ij} = [(\widetilde{e}_i-\widetilde{e}_j)^\top \widetilde{\textbf{L}}^{-1} (\widetilde{e}_i-\widetilde{e}_j)]^{-1/2}$, \myworries{where the inverse of the grounded Laplacian is used in lieu of the pseudoinverse of the standard Laplacian matrix and $\widetilde{e}_i$ and $\widetilde{e}_j$ are respectively $e_i$ and $e_j$ without the $N$-th node.} 
Indeed, the action of the matrix $Q$ can be understood in terms of removing the null-space of $\textbf{L}$ by effectively ``grounding'' the resistor network, i.e., fixing a reference potential~\cite{jadbabaie2004stability,ghosh2008minimizing}.

Young et al.~\cite{young1:2015} now extended the notion of resistance distance by using an alternative characterization of the inverse $\widetilde{\textbf{L}}^{-1}$ that appears in the definition of the effective resistance.
Specifically, they define the \emph{generalized effective resistance distance} between node $k$ and node $j$ by:
\begin{ceqn}
\begin{align}
d_\text{GRD}(k,j)= \sqrt{(e_k-e_j)^\top \textbf{X} (e_k-e_j)}.
\end{align}
\end{ceqn}
Here the matrix $\textbf{X}$ is defined via 
\begin{ceqn}
\begin{align}
\textbf{X} =2\boldsymbol{Q^\top \Sigma Q},
\end{align}
\end{ceqn}
where $\Sigma$ is the solution of the Lyapunov equation:
\begin{ceqn}
\begin{align}
\boldsymbol{\widetilde{L}\Sigma + \Sigma\widetilde{L}^\top= I_{N-1}}
\end{align}
\end{ceqn}
This solution is unique under the assumption that there exists a globally reachable node within the graph.
Note that this formulation reduces to the classical resistance distance in the case of undirected graphs.
For more details, we refer to~\cite{young1:2015,young2:2015}.

\subsubsection{Hitting Time Based Distance (HTD)}
A second type of node-to-node distance metric for directed graphs is the class of hitting time metrics developed by Boyd et al.~\cite{Boyd:2021}. 

Consider a discrete-time Markov chain $(X_{t})_{t \geq 0}$ on the space of the vertices $\mathcal{V} =\{1,...,N\}$ of a strongly connected graph, with initial distribution $\lambda$ and a irreducible transition matrix $\textbf{P}=\textbf{D}^{-1}\textbf{A}$ such that:
\begin{ceqn}
\begin{align}
P(X_{0}=i)=\lambda_{i}\qquad\text{and} \qquad P(X_{t+1}=j|X_{t}=i)=P_{i,j}.  
\end{align}
\end{ceqn}
Let $\pi \in \mathbb{R}^{N}$ be the invariant distribution of the chain, i.e., $\pi \textbf{P}=\pi$. 
For a starting point distributed according to $\lambda$, the \emph{hitting time} of a target vertex $i \in \mathcal{V}$ is the random variable

\begin{ceqn}
\begin{align}
\tau_{i} = \inf\{t \geq 1:X_{t}=i\}.
\end{align}
\end{ceqn}

Following~\cite{Boyd:2021}, we denote the probability that starting in a node $i$ the hitting time of $j$ is less than the time it takes to return back to $i$ by
\begin{ceqn}
\begin{align}
Q_{i,j} := P_{i}[\tau_{j} \leq \tau_{i}].
\end{align}
\end{ceqn}
Based on the matrix $\textbf{Q}$ a normalized hitting time matrix $\textbf{T}^ {(\beta)}$ can be defined in terms of its entries
\begin{equation}
  T^{(\beta)}_{i,j} =
    \begin{cases}
      \frac{\pi^{\beta}_i}{\pi^{1-\beta}_j}Q_{i,j} & \text{$i \neq j$},\\
      1 & \text{otherwise},
    \end{cases}       
\end{equation}
where $\beta$ is a scalar parameter that can be used to adjust the measure
If $\textbf{P}$ is an irreducible stochastic matrix (i.e., the underlying graph is strongly connected, the Hitting Time Distance Matrix for $\beta = (0.5,1]$ can be obtained by:
\begin{equation}
    d_\text{HTD}^{(\beta)}(i,j) = HTD_{i,j}^{(\beta)} \quad \text{where} \quad HTD_{i,j}^{(\beta)}=-\log(T_{i,j}^{(\beta)})
\end{equation}

\noindent\textbf{Remark.} (Not strongly connected graphs).
Note that both distance functions discussed above make assumptions about the global reachability of (all) nodes in the graph.
To make the above distance measures well defined in case these assumptions are not fulfilled, one option is to add a low-rank regularization term, as popularized within the context of the well-known PageRank algorithm~\cite{page1998pagerank,gleich2015pagerank}.

\subsection{Optimal Transport distances for Directed Weighted Graphs}
\myworries{In the following, we will exploit} the geometries defined by the above described node-to-node distances measures $d_\text{GRD}$ \myworries{$d_\text{HTD}^{(\beta)}$ to derive} two optimal transport distances between directed graphs: 1) a Gromov-Wasserstein distance and 2) a Wasserstein distance.

\subsubsection{Gromov-Wasserstein distance for directed graphs}
The intuitive idea underpinning Gromov-Wasserstein distances is to probabilistically map one geometric configuration  (metric-space) onto another, while distorting the distances as weakly as possible.
In the context of graphs these ideas have been explored, e.g., in~\cite{Memoli:2011,Chowdhury:2019}.
To define a Gromow-Wasserstein distance for graphs, we thus need to consider the graph as a collection of points (corresdponding to the nodes), with a (symmetric) metric distance defined between them.
This is precisely for what we will use the above introduced node-to-node distance metrics.

Formally, consider two graphs $\mathcal{G}^{k}$ and $\mathcal{G}^{l}$.
To each graph, we associated the tuple ($\mathcal{C}^{k}$, $\mathbf{p}^{k}$) consisting of a node-to-node distance matrix $\mathcal C^{k}$ and a probability vector $p_k\in\mathbb{R}^N$ which describes the relative ``mass''  we associate to each node in that graph.
Without further available information, we will typically use the uniform distribution $\mathbf{p}^k=\mathbf{1}/N$ as an agnostic choice.

The Gromov-Wasserstein distance between two graphs $\mathcal{G}^k$ and $\mathcal{G}^l$ can then be obtained as the solution to the following minimization problem:
\begin{equation}
d_{\text{GW}}(\mathcal{G}^k,\mathcal{G}^l) = \min_{\mathbf{\Gamma} \in \prod(p_{k},p_{l}) } \quad \sum_{x_1,x_2,y_1,y_2} \mathcal{L}(\mathcal{C}_{x_1,x_2}^k, \mathcal{C}_{y_1,y_2}^l) 
\Gamma_{x_1,y_1} \Gamma_{x_2,y_2} 
\end{equation}
where $\prod(p_{k},p_{l}) = \{\Gamma \in \mathbb{R}^{N\times N}: \Gamma_{ij} \leq 0,  \Gamma\mathbf{1} = \mathbf{p}^{k}, \Gamma^\top\mathbf{1} =\mathbf{p}^{l}\}$ is the set of all possible transport plans with marginal distributions $\mathbf{p}^{k}$, $\mathbf{p}^l$, and $\mathcal L$ is a loss function defined over the distance matrices $\mathcal{C}^{k}$.
Commonly, the loss function $\mathcal L(.,.)$ is simply an (elementwise) $L_{2}$ norm.

\subsubsection{Wasserstein Distance for directed graphs}
To derive a Wasserstein distance for directed graphs, we switch our interpretation of the observed data.
Instead of considering the geometry defined by each graph in our observation set $\lbrace\mathcal{G}^{1},..., \mathcal{G}^{p}\rbrace$ separately, and transporting these geometries such that the distortion is minimized, we now interpret the edge-weights of each weighted graph as a specific realization of a signal supported on the \emph{same underlying space}.
Stated differently, we consider the edge-weights as a distribution supported on the same set of latent edges, and we aim to transport these edge distributions on top of each other optimally.

To this end, we allocate the observed edge weights in a matrix $\varmathbb{P} \in \mathbb{R}^{E \times p}$, whose rows are indexed by the (directed) edges and whose columns by the respective graph, i.e., the column $\varmathbb{P}_{:,k}$ describes the weights of the $k$th graph $\mathcal{G}^k$.

Using this definition, we build a \emph{directed linegraph} $\varmathbb{L}=(\varmathbb{V},\varmathbb{E},\varmathbb{W})$.
The vertex set $\varmathbb{V}$ of the linegraph contains each possible edge $(j,k)$ contained in one of the graphs $\mathcal G^i$.
The edge set $\varmathbb{E}$ of the linegraph is defined as follows: an edge $e^\prime=(u^\prime,v^\prime) \in \varmathbb{E}$ exists if the target of $u^\prime$ is the source of $v^\prime$, i.e., the target node of edge $u^\prime$ in the original graph, is the source node of edge $v^\prime$ in the original graph. 
Note that, to build the line graph, the weights of the interactions are not considered (but are considered as distribution supported on the line-graph).
Finally, the weight $w_{u'v'}$ of edge $(u^\prime,v^\prime)$  in the line graph is simply the proportion of graphs containing both edges $u'$ to $v'$.

Once we have a weighted, directed linegraph $\varmathbb{L}$, we can again associate a distance matrix based on the generalized effective resistance or the hitting time with $\varmathbb{L}$. 
The Wasserstein distance between two graphs $\mathcal{G}^{k}$ and $\mathcal{G}^l$ can then be computed as follow:

\begin{align}
    d_W(\mathcal{G}^{k},\mathcal{G}^l) = \mathop{\arg\min}_{\Gamma \in \Pi(\varmathbb{L})}\quad \langle \Gamma, \mathcal{C}_{\varmathbb{L}} \rangle_F
\end{align}
$\mathcal{C}_{\varmathbb{L}}$ is one of the distance matrices (generalized effective resistance, hitting time distance) defined for directed graphs, as discussed above. 
Here $\Pi(\varmathbb{L}) = \{\Gamma \in \mathbb{R}^{|\varmathbb{V}|\times |\varmathbb{V}|}: \Gamma_{ij} \leq 0,  \Gamma\mathbf{1} = \varmathbb^{P}_{:,k}, \Gamma^\top\mathbf{1} =\varmathbb^{P}_{:,l}\}$ is again the set of all admissible transport matrices, whose marginal distribution is now, however, defined by the respective distribution of edge weights as encoded in the columns of the matrix $\varmathbb{P}$.

\section{Numerical Experiments}
\label{sec:numerical_experiments}
\begin{figure}[tb!]
\begin{minipage}[b]{1.0\linewidth}
\centering
\includegraphics[width=8.5cm]{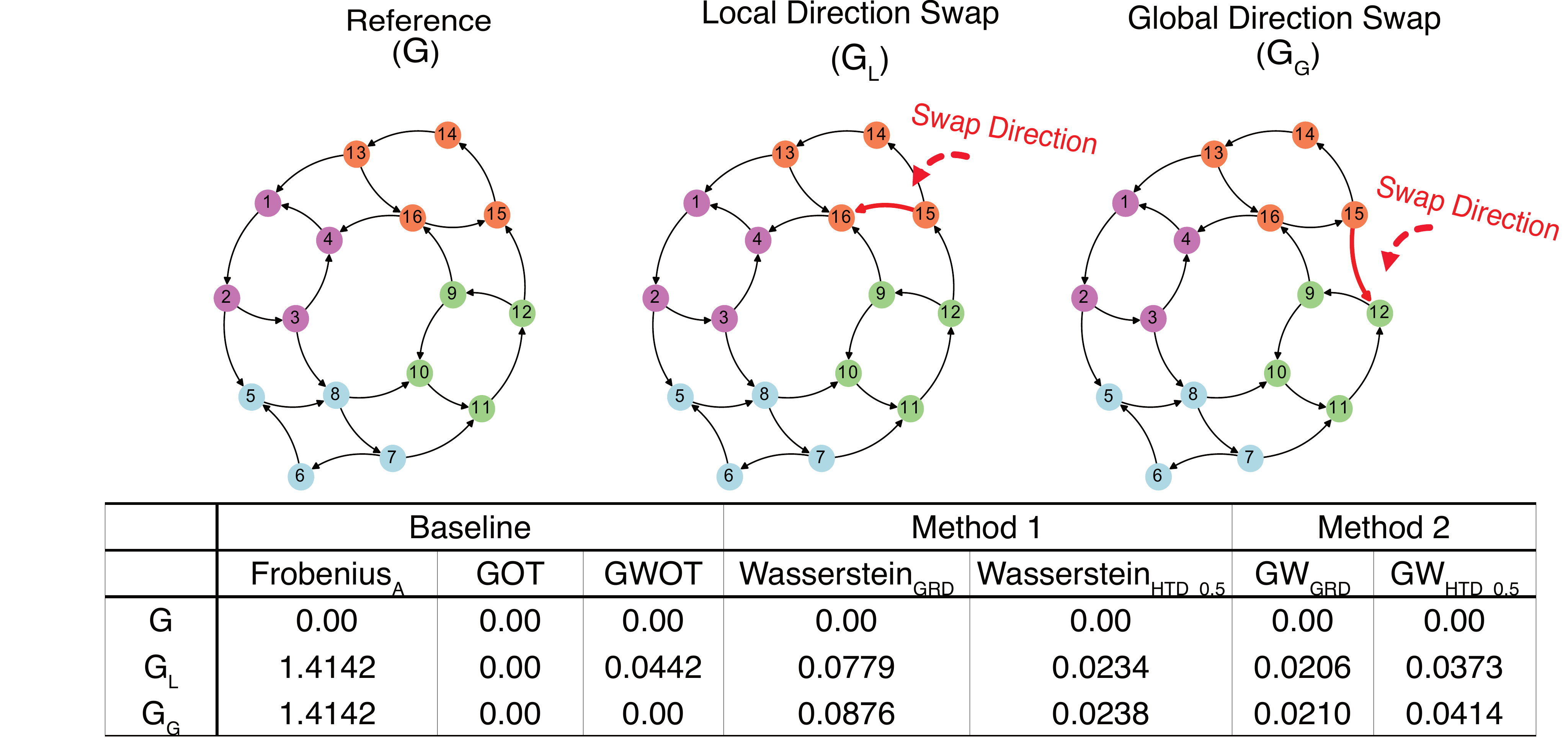}
\caption{\myworries{\textbf{Optimal Transport distance between directed graphs.}}
We consider an illustrative example network, consisting of four directed ``local'' cycles (indicated by node color), who are again connected ``globally'' in a cyclic way.
Left: original reference network (note that the network is strongly connected).
Middle: We perturb the network by swapping the direction of one of the ``local'' edges within one of the four cycles.
Right: We perturb the network by swapping the direction of one of the ``global'' edges connecting the four cycles.
The table below displays the obtained distances considering the following three baselines: the Frobenius norm of the difference between the adjacency matrices; the optimal transport-based GOT distance~\cite{Maretic:2019}, when considering the graph as undirected, and the optimal transport-based GWOT distance~\cite{Chowdhury:2019} (again considering the graph as undirected). We contrast these with the results obtained via our Gromow-Wasserstein and Wasserstein (earth movers distance) formulation based on both generalized effective resistance and hitting time distance. Note that only our distance metrics that account for the directionality of the graph edges can distinguish these different cases.}
\label{fig:distance}
\end{minipage}
\end{figure}
\myworries{
\begin{table*}[!tb]
    \centering
    \caption{Adjusted Rand (AR) measuring the accuracy of DWG clustering. Values in bold indicate the outperforming method.}
    \resizebox{\textwidth}{!}{
        \begin{tabular}{lrr|rrrrrrrr}
        \toprule
        Dataset & no. Classes & $|V|_{max}$ & PCA(P) & correlation(P) & $GW_{HTD^{0.5}}$ & $GW_{GRD}$ & $GW_{HTD^1}$ & $Wasserstein_{GRD}$ & $Wasserstein_{HTD^1}$ & $Wasserstein_{HTD^{0.5}}$ \\\midrule
        Pancreas Cancer(n=35) & 2 & 10 & -0.047014 & -0.033237	 & 0.109974 & 0.109974
 & 0.109974	 &\textbf{ 0.884777 }& \textbf{0.884777} & \textbf{0.884777} \\
        Heart Myocardial Infarction(n=20) & 2 & 33 & -0.041644	 & 0.312917	 & 0.312917 & 0.312917	 & 0.312917 & \textbf{0.455679} & \textbf{0.455679} & \textbf{0.455679}\\\bottomrule
        \end{tabular}
    }
    \label{tab:Ari}
\end{table*}
}
\subsection{Experiments on synthetic graphs: an illustrative example}

To illustrate the behavior of our distance measures for directed weighted graphs, we consider a synthetic example in which we connected four cyclic graphs with four nodes each in a cyclic fashion --- see~\Cref{fig:distance} for an illustration.
We now consider two different perturbations of these graphs and assess how far these changes can be detected.
Both of these perturbations consist merely of a flip of the orientation of one of the edges, meaning that when the graph is considered undirected there is no apparent change in the graph structure.
The first perturbation we consider is a flip in orientation in one of the ``local'' edges within one of the cyclic four-node-graphs (cf.~\Cref{fig:distance}).
The second perturbation we consider is a flip in the orientation of one of the ``global'' edges that connect two of the cyclic graphs (cf.~\Cref{fig:distance}).

The distances between these perturbed graphs, relative to the original graph, are displayed in the table shown in~\Cref{fig:distance}.
We see that both of our distance metrics for directed graphs capture a difference between the two scenarios, irrespective of what directed node-to-node metric we use for them.
In fact, in most cases, the ``global'' edge flip leads to a larger distance.
If we consider the directed graph as a flow circulation pattern, it can be argued that, indeed, this single ``global'' flip leads to a larger overall perturbation in the overall flow pattern, in contrast to flipping a ``local'' edge which leads only to a reversal of the flow within one of the cycles.

We also compare our distances, to standard distance metrics for undirected graphs as baselines, i.e., the Frobenius norm,  GOT~\cite{Maretic:2019}. Furthermore, a previously proposed method for comparing directed graphs, GWOT~\cite{Chowdhury:2019}, was also considered in our benchmark.
Interestingly, this approach also fails to distinguish the graph with the global edge flip from the original graph.
From the evaluated distances, the Frobenius norm identifies the two perturbed graphs to be equally far apart from our original graph.
As there is no difference in the undirected graph structure, the GOT distance assigns a zero distance between all graphs. 
Overall, we see that accounting for directions in the edges is thus important to obtain meaningful distances between the graphs.

\subsection{Case Study: single-cell RNAseq derived cell-cell interactions disease clustering}

To illustrate the utility of our graph distances, we further consider a real-world analysis task, namely the comparison of cell-cell communication networks.
Specifically, we consider a scenario in which we are given a patient cohort with $p$ patients, from which we generate a set of cell-cell communication networks $\mathcal{\textbf{G}}=\{\mathcal{G}^{1},..., \mathcal{G}^{p}\}$ using the ligand-receptor analysis method CrossTalkeR~\cite{Nagai:2021}, for more detailed description of the networks we refer to~\cite{Nagai:2021}. 

Our question here is whether the comparison of cell-cell networks can be used to delineate the disease stage or sub-types in these patients. 
To construct our set of graphs, we retrieved two single-cell RNA sequencing datasets that are publicly available. These data have 20-35 samples, which are annotated regarding the health status of patients.  
Next, we compute the graph-to-graph distances using our introduced distance measures.

To reveal how far we can find a partition of the patient cohorts, we perform $k$-means with the estimated distance matrix as input. For simplicity, we set $k$ to the (true) number of classes in the data, and use the Adjusted Rand index (ARI)~\cite{Hubert:1985} between the true label and clustering result to evaluate the performance of the clustering result.
We also include a few baseline methods. In the first baseline, we would like to compare our methods with a simplistic way of obtaining a distance using the graph set $\mathbf{G}$. Here, the matrix $\varmathbb{P}$ is used as an input for a Principal Component Analysis, which can be obtained by computing an SVD of the centered $\varmathbb{P}$ matrix $\tilde{\varmathbb{P}} = U\Sigma V^\top$ 
Next, the principal components matrix $U_{p \times (p-1)}$ are used to compute the pairwise Euclidean distance between the samples. 

In the second baseline, we attempt to compare our method(directed) with an undirected cost function. For that, we compute the correlation distance $dCor(u,v)$, between the pair of rows from $\varmathbb{P}$.
\begin{equation}
    dCor(u,v) = 1 - \frac{\langle(u - \bar{u}),(v - \bar{v})\rangle}
                  {{\|(u - \bar{u})\|}_2 {\|(v - \bar{v})\|}_2},
\end{equation}
where $\bar{u}$ and $\bar{v}$ denote the average of the vectors $u$ and $v$, respectively.
The distance matrix resulting from that process is then used as a cost function for the Wasserstein approach described above.

We display the results of these experiments in Table~\ref{tab:Ari}.
As can be seen, our optimal transport-based distances outperform the baseline methods (correlation and PCA), disregarding edge directions. 
Interestingly, we find that in this task, the Wasserstein-based construction outperforms the Gromow-Wasserstein formulation.
In contrast, the node-to-node distance metric used to induce a geometry for the directed graphs appears to play a far less significant role. 
These results indicate that the Wasserstein formulation might be better suited for sparse and noisy DWG obtained from cell-cell communication networks from single-cell RNA sequencing. 
A potential explanation may be that the Wasserstein formulation is adaptive with respect to the whole ensemble of graphs considered, due to the specific construction of the linegraph. 

\section{Conclusions}

We presented two OT-based distance measures to compare directed graphs, based upon node-to-node distance metric for directed graphs. 
Our results highlight the importance of considering edge directions when comparing graphs.
Indeed, as our example in \Cref{fig:distance} shows our approach can capture essential features related to the direction of each interaction in opposition to our baseline models. 

We further provided a case study of cell-cell communication networks as a  real word application of the proposed methods. 
Here our methods enabled the identification of latent disease stages and potentially disease-related interaction groups. 

In future work, we intend to improve the robustness of our structural comparison, e.g., by adding other synthetic directed weighted graph models. 
We will also aim to characterize in more detail the effects of noise and sparseness in our methods as well as explore approaches using the computed transport maps for explainability. For instance, in the context of our example application, we may want to highlight which cell-cell pairs provide the core contributions to the observed differences between healthy and disease states.

\bibliographystyle{IEEEbib}
\bibliography{refs}

\end{document}